# INVESTIGATIONS ON END-TO-END AUDIOVISUAL FUSION

*Michael Wand[1], Ngoc Thang Vu[2], Jürgen Schmidhuber[1]*

[1]Istituto Dalle Molle di studi sull'Intelligenza Artificiale (IDSIA),
USI & SUPSI, Manno-Lugano, Switzerland
[2]Institute for Natural Language Processing (IMS), University of Stuttgart, Germany

## ABSTRACT

Audiovisual speech recognition (AVSR) is a method to alleviate the adverse effect of noise in the acoustic signal. Leveraging recent developments in deep neural network-based speech recognition, we present an AVSR neural network architecture which is trained end-to-end, without the need to separately model the process of decision fusion as in conventional (e.g. HMM-based) systems. The fusion system outperforms single-modality recognition under all noise conditions. Investigation of the *saliency* of the input features shows that the neural network automatically adapts to different noise levels in the acoustic signal.

## 1. INTRODUCTION

Even though automatic speech recognition has seen major technological and performance improvements in the past decade, its accuracy still degrades under adverse acoustic conditions, i.e. when noise is present in the speech signal. *Audiovisual* speech recognition (AVSR), where video images of the speaker's face are taken as additional input, can be used to reduce the adverse effect of acoustic noise [1, 2, 3].

In such a *multimodal* system, it becomes an important question how to combine these modalities. This is particularly true when their reliabilities differ, which is true for AVSR and is known to require an adaptive weighting of the contribution of the input modalities [3, 4]. In this study we present an LSTM [5] neural network for audiovisual speech recognition at different levels of acoustic noise which is trained end-to-end, without an explicit model for the reliability of the input streams. We show that the model remains robust under several different input noise levels, and an investigation of the *saliency* of the input features [6] shows that the neural network tends to automatically adjust to the level of input noise by adjusting the relevance of the different input modalities.

The first author was supported by the H2020 project INPUT (grant #687795). This work used computational resources from the Swiss National Supercomputing Centre (CSCS) under project ID d74.

## 2. RELATED WORK

Lipreading to augment conventional acoustic speech recognition has first been proposed by Petajan in his PhD thesis [1]. Since then, diverse methods for visual-only and audiovisual speech recognition have been described, often using complex features from the computer vision domain and HMMs or related methods for sequence modeling [2, 3].

To the best of our knowledge, the first application of a neural network to sequential audiovisual data was described by Stork and colleagues [7]. Since then, major developments in machine learning and speech recognition have regularly found their way into the field, including network pretraining using RBMs [8] and feature extraction with convolutional networks [9]. End-to-end training of a complete audiovisual speech processing pipeline has been introduced in 2016 [10], and more recently a recognizer trained on TV broadcast data was presented, making a step towards large-scale automated data collection [11]. Audiovisual fusion with neural networks has been reported in prior work [8, 12, 13, 14], however none of those studies use end-to-end training.

Optimal weighting of the input modalities is crucial for achieving best results [3]. This issue has recently been investigated for both conventional systems [4, 15] and for hybrid DNN-HMM systems [13, 14]. Dynamic estimation of stream weights improves performance, however this requires a separate computational step.

Visual and audiovisual speech recognition form part of a large body of work on *biosignal-based* speech processing, i.e. speech capturing, recognition, synthesis, and understanding using relevant biophysiological signals beyond acoustics. Besides noise robustness, such systems have been proposed for speech prostheses, confidential and private communication, and assistance in speech learning and therapy [16, 17].

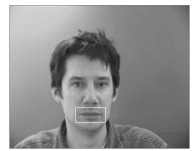

**Fig. 1**: Example frame from GRID with extracted mouth area

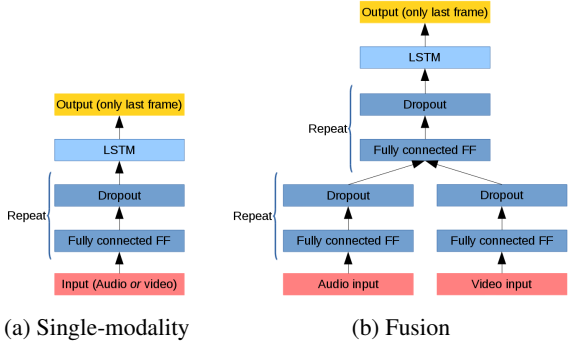

(a) Single-modality  (b) Fusion

**Fig. 2**: Neural network model structures

## 3. DATA CORPUS

We use the GRID audiovisual corpus [18], which consists of video and audio recordings of 34 speakers each saying 1000 sentences, for a total of 28 hours. All sentences contain six words and are of the form *command(4) + color(4) + preposition(4) + letter(25) + digit(10) + adverb(4)*, where the number of alternatives words is given in parentheses. There are 51 different words, note that the corpus is constructed so that the probability of each word is independent of its neighboring words. We use the provided word-level segmentation, obtaining our dataset of 6000 single words per speaker.

To reduce the adverse effect of speaker variation, all experiments in this paper are *speaker-dependent*. The data of each speaker is subdivided into training, validation, and test set as in our prior publications [10, 19], so that validation and test set each contain five samples per word. Furthermore, the entire dataset is subdivided into the *development* speakers #1 – #20 and the *evaluation* speakers #22 - #34[1].

**Audio** data is preprocessed with OpenSMILE [20]: The waveform is windowed with 20ms length and 10ms shift, for each window 27 log Mel-scale energy bands are computed. Noisy audio is created by superimposing Babble noises from the *Freesound* database [21] to the clean audio at {-5dB, 0dB, 5dB} SNR, using the acoustic simulator presented in ref. [22].

For **video** data, raw pixel features are used. The mouth ROI is extracted from the full-face videos as follows: First, facial landmarks are detected with the DLib facial landmark detector [23], using the sample implementation by A. Rosebrock[2]. The area covered by the mouth landmarks #49 - #68 is enlarged by 10 pixels to the left and right, resized to 80x40 pixels, and considered the ROI. We visually confirmed that this gives reasonable mouth area estimates; figure 1 shows an example frame indicating the computed ROI.

Finally, all videos are converted into grayscale. In the fusion experiments, the video stream is upsampled by a factor of 4 to obtain 100 frames/second as for the audio.

---
[1]Speaker 21 is excluded because no videos are provided for this speaker.
[2]https://www.pyimagesearch.com/2017/04/03/facial-landmarks-dlib-opencv-python

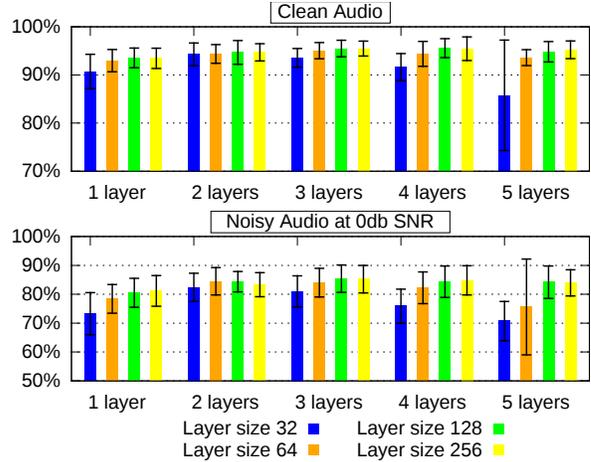

**Fig. 3**: Recognition results on **audio** data for different network architectures, averaged over the development speakers

## 4. SPEECH RECOGNITION EXPERIMENTS

### 4.1. Single-modality baseline

We first establish accuracy baselines using *only* acoustic or *only* visual data, intending to cover a reasonable range of network parameters and to report optimal baseline results. In order to keep the complexity of this task under control, we impose the following constraints: As in prior work [10, 19], the network is composed of a stack of fully connected layers with tanh nonlinearity, each followed by Dropout with 50% probability, a single LSTM layer with tanh nonlinearity, and a softmax layer with 51 word output neurons (see figure 2a). All layers (including the LSTM layer) have the same number of neurons, convolutional layers are not used [10].

The loss is given by the multiclass cross entropy, summed over a minibatch of 64 words. The training error is backpropagated only from the last frame of each word; likewise during testing, only the prediction on the last frame is used. Training uses gradient descent with a learn rate of 0.001 and a momentum of 0.5. (We also experimented with other metaparameters and found these settings optimal.) All experiments were performed using Tensorflow [24].

Figure 3 shows the accuracy of the *audio-only* recognizer for clean audio data and for audio with 0dB babble noise. Accuracy decreases with smaller layers [10]; in a range between 2 and 4 layers (including the LSTM layer) of 64 – 256 neurons, the differences between the architectures are minimal. The best accuracy is 95.6% for clean audio, using four 128-neuron layers, for 0dB noise, the best result is 85.4% accuracy with three 128-neuron layers. Since the difference is marginal, we use the latter architecture as a common baseline.

When training on data from *all four* noise levels (including clean audio), the accuracy improves for noisy audio, but decreases for clean audio, see table 1. We assume that this effect is due to the regularizing properties of input noise.

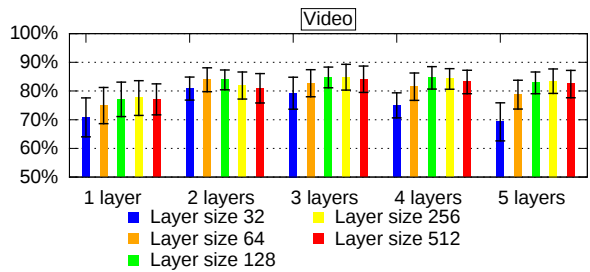

**Fig. 4**: Recognition results on **video** data for different network architectures, averaged over the development speakers

Figure 4 shows word accuracies for *video-only* recognition. The trend is similar to the audio case, the best result of 84.8% accuracy is achieved with three layers with 256 neurons each. Again, architectures within a reasonable parameter range perform similarly, thus to have a common architecture for all experiments, we set the video baseline at three layers with 128 neurons each (84.7% accuracy); cp. table 1.

### 4.2. Fusion experiments on single noise levels

The fusion network architecture is shown in figure 2b: Audio and video input is fed into separate sub-networks whose outputs are merged and form the input of a joint network (cp. [8, 12]). Early fusion, i.e. fusion at the feature level, is contained as a special case, namely when the single-modality sub-networks have zero layers. The whole architecture is trained end-to-end as described in section 4.1.

We searched the optimal fusion network architecture using the constraint that the two modalities use the same number of separate layers, and that all layers have the same number of neurons; this is a reasonable simplification given the results in section 4.1. We found best results with a layer size of 128 and "2+2" separate+joint layers, as displayed in figure 5. Audio-visual fusion clearly improves over the baseline, and the result carries over to systems trained at all noise levels; see table 1. Note in particular that early fusion (i.e. "0+$x$" architectures) gives substantially worse results than fusion at higher levels in the neural network, confirming the related observation in ref. [14]. This may be due to the vastly different number of input features between the audio and video inputs, indeed a PCA dimensionality reduction of the video data [25, 10] improves the accuracy with early fusion.

| Training on single noise level | | | | |
|---|---|---|---|---|
| Noise | Video only | Audio only | Fusion | Rel. Imp. |
| no noise | | 95.5% ± 1.7% | 95.9% ± 2.1% | 8.8% |
| 5dB | 84.7% ± 3.6% | 91.3% ± 3.1% | 93.9% ± 2.7% | 29.9% |
| 0dB | | 85.4% ± 4.7% | 91.6% ± 3.2% | 43.8% |
| -5dB | | 75.8% ± 6.5% | 87.9% ± 3.3% | 50.0% |
| Training on all noise levels | | | | |
| Noise | Video only | Audio only | Fusion | Rel. Imp. |
| no noise | | 93.9% ± 2.4% | 94.0% ± 2.3% | 1.6% |
| 5dB | 84.7% ± 3.6% | 92.3% ± 2.9% | 93.4% ± 2.5% | 14.3% |
| 0dB | | 88.4% ± 4.4% | 92.6% ± 2.4% | 36.2% |
| -5dB | | 78.1% ± 6.7% | 90.4% ± 3.1% | 56.2% |

**Table 1**: Accuracies at different noise levels, for the best setup (three layers with 128 neurons each). Improvement is given as error reduction compared to the audio-only baseline. Averaged over development speakers.

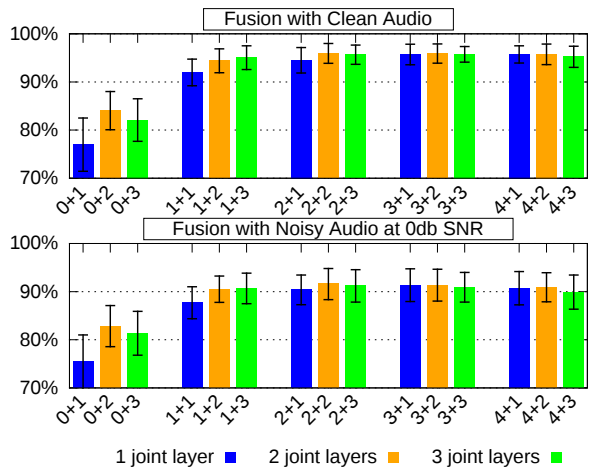

**Fig. 5**: Recognition results of the **audio-visual fusion** experiments for different network architectures, averaged over the development speakers. A network architecture is given as "$x+y$", where $x$ stands for the number of layers separated by modality, and $y$ stands for the number of joint layers (compare figure 2b). All layers have 128 neurons.

## 5. SALIENCY

We see from table 1 that the accuracies of fusion systems are bounded from below by the accuracies of the single modalities. Intuitively, this means that if one modality performs worse than the other, the neural network should give a higher weight to the other.

Class *saliency* maps [6], which indicate the influence of parts of an input sample on the network output, offer a method to investigate this assumption. The idea is to *linearize* the network around the output neuron corresponding to the associated word label by computing the derivative of the output with respect to the input: Assume $I_A$ and $I_V$ are the audio and video input (i.e. videos of arbitrary length). The trained network computes the logistic class score $S_C(I_A, I_V)$ for any word class $C$. $S_C$ is a complicated nonlinear function, which can be approximated in first order by taking the derivatives

$$M_A = \frac{\partial S_C}{\partial I_A} \qquad \text{and} \qquad M_V = \frac{\partial S_C}{\partial I_V}.$$

| | Training on single noise level | | Training on all noise levels | |
|---|---|---|---|---|
| Test Noise | Audio | Video | Audio | Video |
| no noise | 1.50 ± 0.49 | 4.00 ± 0.88 | 1.49 ± 0.46 | 5.45 ± 1.09 |
| 5dB | 1.29 ± 0.35 | 4.51 ± 0.91 | 1.48 ± 0.46 | 5.51 ± 1.14 |
| 0dB | 1.26 ± 0.35 | 5.26 ± 1.09 | 1.51 ± 0.49 | 5.73 ± 1.13 |
| -5dB | 1.19 ± 0.33 | 6.05 ± 1.04 | 1.61 ± 0.48 | 6.20 ± 1.13 |

**Table 2**: Saliencies in the fusion system for audio and video modalitites. Averaged over development speakers.

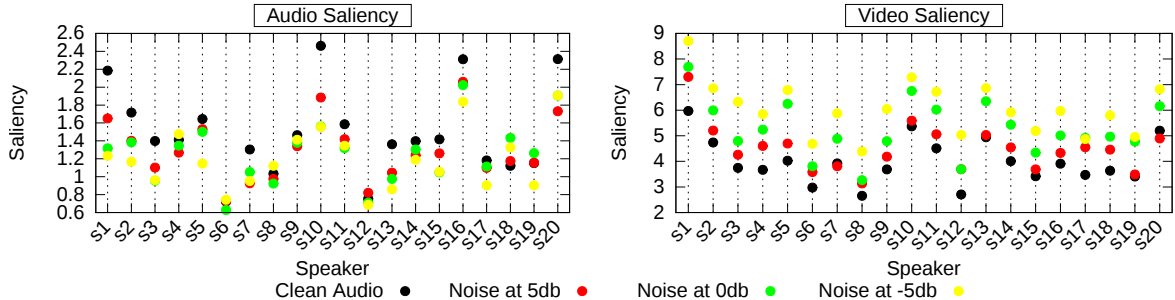

**Fig. 6**: Speaker breakdown of saliencies in the fusion system. Systems were trained on single noise levels.

The saliency is given by taking the absolute value of $M_A$ and $M_V$. Ref. [6] considers the saliencies per input pixel (for static images), we sum the saliencies of each input vector and take the average over all frames of a sequence, and over all sequences of the evaluation set per speaker, thus measuring the relevance of entire modalities.

Figure 6 and the left part of table 2 show the saliencies for audio and video, for fusion systems trained at *single* noise levels. The result in particular for the video modality shows a striking regularity: Almost without exception, the video saliency follows the audio noise level, i.e. more acoustic noise implies a higher video saliency. The behavior of the audio saliency is less clear, but one can see that it is frequently higher for lower noise levels, as assumed. Irregularities might point to specific issues of visual speech recognition, for example, the low distinguishability of certain word pairs [10] might force the system to resort to audio data even in the presence of noise. Performing the saliency computation on fusion systems trained on *all* noise levels shows a similar pattern as before for the video saliencies, see the right part of table 2. The acoustic saliencies, however, behave differently.

## 6. EVALUATION AND CONCLUSION

In this section we summarize the main results of this paper, perform statistical tests on the *evaluation* speakers where appropiate, and lay out directions for future work. Statistical tests are always performed across speakers, using the one-tailed t-test with paired samples at a confidence level of 95%.

We validate the two main results of this paper: 1) modality fusion improves over the audio-only baseline (at different noise levels), 2) considering two fusion systems, the one with the higher noise level exhibits the higher video saliency. Table 3 shows the average accuracies of single-modality and fusion systems on the evaluation speakers; all improvements on *noisy* audio data are significant. Table 4 shows the audio and video saliencies at different noise levels. Since we did not obtain a consistent result on the audio saliency, we only verify the hypothesis that between neighboring noise levels (i.e. between clean audio and 5dB noise, 5dB noise and 0dB noise, etc.), the video saliency increases. This is indeed always the case, all differences are significant.

This shows that the fusion network not only makes use of both input modalities, but also adapts to the present noise level. Future research is required to understand whether this behavior remains robust when different types of noise, or possibly even noisy video data, is used. Furthermore, the irregular audio saliencies point to the fact that the way the network adapts to the underlying signal *and* to the classification task may be more complicated than shown here, in particular regarding the different confusability patterns between modalities.

| Training on single noise level | | | | |
|---|---|---|---|---|
| Noise | Video only | Audio only | Fusion | Rel. Imp. |
| no noise | | 95.3% ± 2.0% | 96.2% ± 1.7% | 19.1% |
| 5dB | 83.0% ± 4.5% | 90.7% ± 3.2% | 93.5% ± 2.3% | 30.1%∗ |
| 0dB | | 85.2% ± 4.1% | 90.9% ± 3.0% | 38.5%∗ |
| -5dB | | 74.4% ± 6.1% | 86.9% ± 3.8% | 48.8%∗ |
| Training on all noise levels | | | | |
| Noise | Video only | Audio only | Fusion | Rel. Imp. |
| no noise | | 93.8% ± 2.2% | 94.4% ± 2.5% | 9.7% |
| 5dB | 83.0% ± 4.5% | 91.5% ± 2.7% | 93.8% ± 2.7% | 27.1%∗ |
| 0dB | | 87.3% ± 4.1% | 92.9% ± 3.0% | 44.1%∗ |
| -5dB | | 78.2% ± 5.5% | 90.3% ± 3.5% | 55.5%∗ |

**Table 3**: Accuracies on the *evaluation* speakers, for the best setup (three layers with 128 neurons each). Improvement is given as error reduction compared to the audio-only baseline; significant improvements are marked with ∗.

| | Training on single noise level | | Training on all noise levels | |
|---|---|---|---|---|
| Test Noise | Audio | Video | Audio | Video |
| no noise | 1.82 ± 0.71 | 3.92 ± 0.68 | 1.87 ± 0.67 | 5.37 ± 1.09 |
| 5dB | 1.60 ± 0.59 | 4.66 ± 0.82∗ | 1.89 ± 0.64 | 5.54 ± 1.10∗ |
| 0dB | 1.46 ± 0.41 | 5.13 ± 0.86∗ | 1.88 ± 0.63 | 5.65 ± 1.12∗ |
| -5dB | 1.42 ± 0.52 | 5.93 ± 1.12∗ | 2.05 ± 0.63 | 6.27 ± 1.11∗ |

**Table 4**: Saliencies in the fusion system for audio and video modalitites. Averaged over evaluation speakers. Significant differences of the video saliency to next lower noise level marked by ∗.